\documentclass[lettersize,journal]{IEEEtran}
\usepackage{amsmath,amsfonts}
\usepackage{algorithmic}
\usepackage{algorithm}
\usepackage{array}
\usepackage{multirow}
\usepackage[caption=false,font=normalsize,labelfont=sf,textfont=sf]{subfig}
\usepackage{textcomp}
\usepackage{stfloats}
\usepackage{url}
\usepackage{verbatim}
\usepackage{graphicx}
\usepackage{cite}
\begin{document}

\title{MapSAM: Adapting Segment Anything Model for Automated Feature Detection in Historical Maps}

\author{Xue Xia,~\IEEEmembership{Graduate Student Member,~IEEE}, Daiwei Zhang, Wenxuan Song, Wei Huang, Lorenz Hurni
        % <-this % stops a space
\thanks{This work was supported by the Swiss National Science Foundation as part of the EMPHASES Project [Grant Number: 200021\_192018]. (\textit{Corresponding author: Xue Xia.})}%
\thanks{Xue Xia and Lorenz Hurni are with the Institute of Cartography and Geoinformation, ETH Zurich, 8093 Zurich, Switzerland (e-mail: xiaxue@ethz.ch).}% <-this % stops a space
\thanks{Daiwei Zhang and Wenxuan Song are with the Department of Computer Science, ETH Zurich, 8006 Zurich, Switzerland.}%
\thanks{Wei Huang is with the Chair of Data Science in Earth Observation, Technical University of Munich, 80333 Munich, Germany.}}

\markboth{preprint under review}%
{XIA \MakeLowercase{\textit{et al.}}: MapSAM: Adapting Segment Anything Model for Automated Feature Detection in Historical Maps}


\maketitle

\begin{abstract}
Automated feature detection in historical maps can significantly accelerate the reconstruction of the geospatial past. However, this process is often constrained by the time-consuming task of manually digitizing sufficient high-quality training data. The emergence of visual foundation models, such as the Segment Anything Model (SAM), offers a promising solution due to their remarkable generalization capabilities and rapid adaptation to new data distributions. Despite this, directly applying SAM in a zero-shot manner to historical map segmentation poses significant challenges, including poor recognition of certain geospatial features and a reliance on input prompts, which limits its ability to be fully automated. To address these challenges, we introduce MapSAM, a parameter-efficient fine-tuning strategy that adapts SAM into a prompt-free and versatile solution for various downstream historical map segmentation tasks.  Specifically, we employ Weight-Decomposed Low-Rank Adaptation (DoRA) to integrate domain-specific knowledge into the image encoder. Additionally, we develop an automatic prompt generation process, eliminating the need for manual input. We further enhance the positional prompt in SAM, transforming it into a higher-level positional-semantic prompt, and modify the cross-attention mechanism in the mask decoder with masked attention for more effective feature aggregation. The proposed MapSAM framework demonstrates promising performance across two distinct historical map segmentation tasks: one focused on linear features and the other on areal features. Experimental results show that it adapts well to various features, even when fine-tuned with extremely limited data (e.g. 10 shots). The code will be available at https://github.com/Xue-Xia/MapSAM.  
\end{abstract}


\section{Introduction}
\IEEEPARstart{F}{rom} the 18th century onwards, the rapid development of cartography and geodesy led to the massive production of topographic maps at various scales \cite{Chazalon2021}. For centuries, these maps have provided rich, detailed, and often geometrically accurate representations of numerous geographic entities. They serve as valuable resources for studying the evolution of urbanization and infrastructure, understanding geographic and anthropogenic changes over time, and exploring socio-ecological system dynamics \cite{Heitzler2019, Xia2022}. While information in raw scanned historical maps cannot be directly utilized, efficiently and effectively extracting features from these maps is crucial for ensuring accessibility and enabling spatio-temporal analysis for researchers and the public.

The process of detecting the precise locations of geographic features from historical maps and assigning them semantic labels is known as historical map segmentation \cite{Chiang2020}. Current approaches to historical map segmentation are developed using traditional Convolutional Neural Networks (CNN) or Vision Transformers (ViT) \cite{Dosovitskiy2020}, which require large volumes of high-quality labeled training data \cite{Heitzler2020, Jiao2022, Xia2024}. However, the manual labeling process for map data is extremely labor-intensive, demanding significant time and effort to ensure accuracy and consistency across the dataset.

Recently, The rise of vision foundation models has introduced a new research paradigm within the realm of image segmentation \cite{Chen2023}. Departing from the conventional approach of training domain-specific CNN or ViT models with extensive training data, foundation models, with their remarkable generalization capability, can quickly adapt to new downstream tasks with only low-resource fine-tuning or even zero-shot learning \cite{Zhang2023, Roth2024, Zhang2023a}. The Segment Anything Model (SAM) \cite{Kirillov2023} is one of the first vision foundation models for image segmentation. The core design of SAM lies with “Promptable Segmentation”, a concept where a manually crafted prompt (e.g., points, boxes, masks, text) serves as input, yielding the desired segmentation mask as output. 

\begin{figure}[!t]
\centering
\includegraphics[width=\columnwidth]{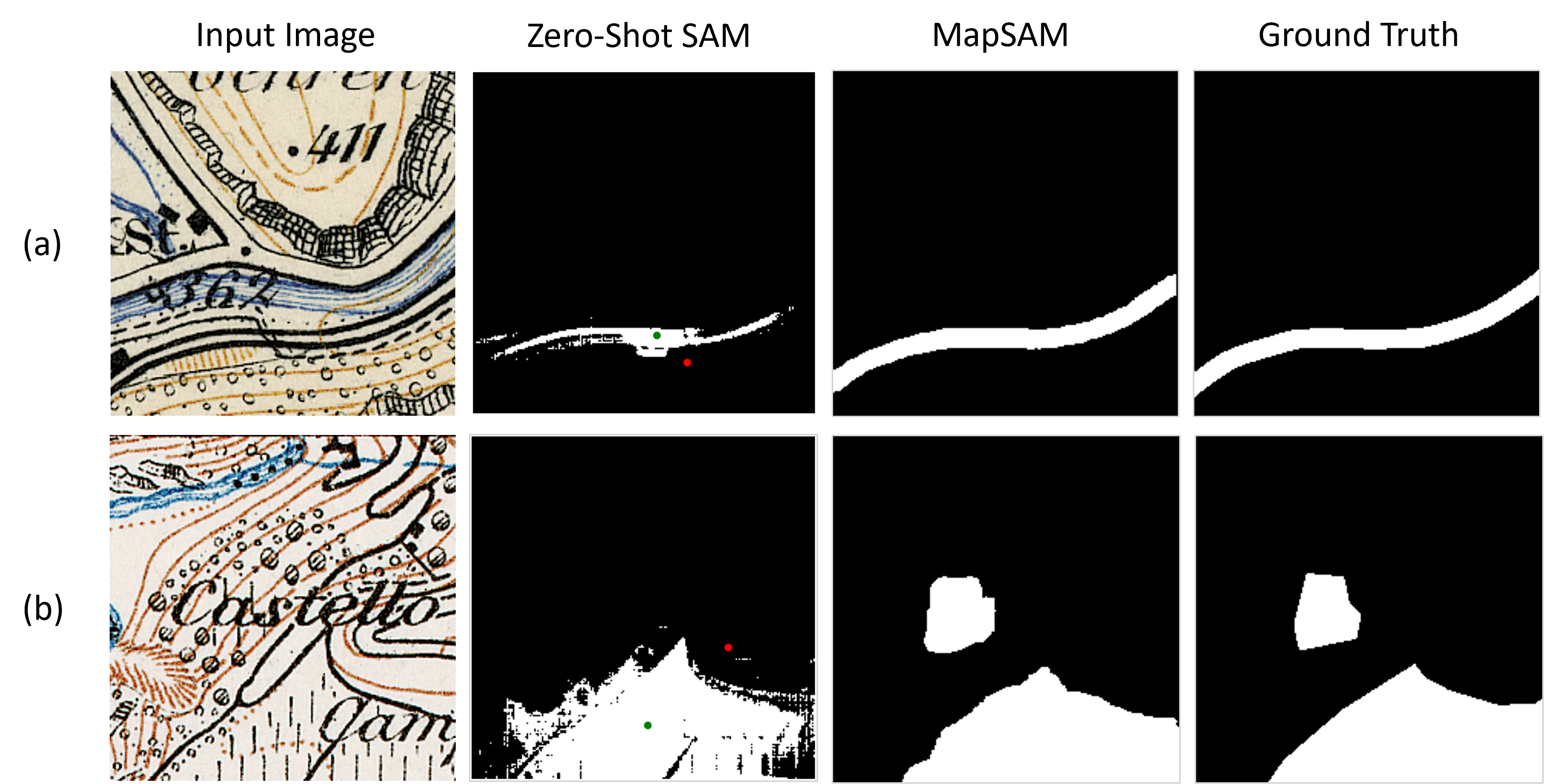}
\caption{Examples of using zero-shot SAM and MapSAM for feature segmentation in historical maps. Rows (a) and (b) depict the segmentation of railways (two thick parallel lines) and vineyards (a set of vertical strokes), respectively. Zero-shot SAM requires manual prompts—green and red points representing positive and negative point prompts, respectively—but fails to delineate clear boundaries for the target objects effectively. In contrast, MapSAM eliminates the need for manual intervention and significantly improves segmentation accuracy.}
\label{fig1}
\end{figure}

Fig. \ref{fig1} showcases examples of using zero-shot SAM to segment features in historical maps. When directly applied to historical map segmentation, vanilla SAM encounters two major shortcomings: 1) It struggles to generalize effectively across various feature detection tasks in historical maps, indicating the need for incorporating domain-specific knowledge to enhance its performance; 2) It relies on human intervention to provide prompts, highlighting the desirability of a fully automatic solution. To overcome these challenges, we introduce MapSAM, an end-to-end framework designed to efficiently adapt SAM to the historical map domain.

MapSAM adopts parameter-efficient fine-tuning (PEFT) and automatic prompting to customize SAM for historical map segmentation. Specifically, we freeze the heavyweight image encoder, and insert learnable DoRA (Weight-Decomposed Low-Rank Adaptation) \cite{Liu2024} layers to acquire domain-specific knowledge for map feature extraction. DoRA does not alter the model architecture. Instead, it serves as a low-rank decomposition of the pre-trained weight matrix, enabling fine-tuning with only a minimal number of parameters. This approach allows for efficient adaptation of the pre-trained model to the target domain, even with limited resources for fine-tuning.  Besides, to eliminate the reliance on human intervention for input prompts, we introduce an auto-prompt generator module‒a specially designed CNN that creates a coarse mask by fusing image embeddings from different encoder layers. Based on the confidence scores in the coarse mask, two points are selected as positive and negative location priors, serving as point prompts to be fed into the prompt encoder. Additionally, inspired by PerSAM \cite{Zhang2023a}, providing low-level positional prompts (e.g., the coordinates of a point) along with high-level target semantics offers the decoder additional visual cues, thus enhancing the segmentation result. Therefore, we extract the embedding of the target object using the coarse mask and fuse it with the original prompt tokens to form the final positional-semantic prompt tokens. These prompt tokens interact with the image embedding within the mask decoder to generate the final segmentation mask. However, Transformer-based models are often claimed to have slow convergence due to global context processing in the cross-attention layers of the Transformer decoder \cite{Gao2021, Sun2021}. To address this, Mask2Former \cite{Cheng2022} proposes masked attention, which constrains attention within the foreground target regions for intensive feature aggregation, achieving better convergence and results. Motivated by it, we introduce masked attention in MapSAM by leveraging the coarse mask to compel the prompt tokens to attend to localized object regions.  

With the aforementioned designs, we apply MapSAM to two customized historical map datasets: one featuring linear features for railway detection and the other featuring areal objects for vineyard detection. Both demonstrate favorable segmentation performance. We summarize the contributions of our paper as follows:
\begin{enumerate}
\item We pioneer the adaptation of the foundation model, SAM, for historical map segmentation. The proposed MapSAM is versatile and prompt-free, capable of automatically detecting both linear and areal features without the need for modifying the model architecture.
\item By incorporating advanced techniques tailored to historical map data—such as low-rank-based fine-tuning, auto-prompt generation, multi-layer feature fusion, positive-negative location prior selection, and masked attention—our framework achieves highly competitive results against domain-specific CNN models like U-Net \cite{Ronneberger2015} and current state-of-the-art SAM-based models like SAMed \cite{Zhang2023} and few-shot SAM \cite{Wu2024}.
\item Since the trainable parameters of our method are lightweight, MapSAM also exhibits outstanding performance in a few-shot learning setting, demonstrating a strong capability to generalize from a minimal amount of training data.
\item There is a significant shortage of widely accessible historical map datasets for researchers to evaluate their techniques. To promote the advancement of automatic feature detection in the historical map domain, we will make our railway and vineyard datasets publicly available at https://doi.org/10.3929/ethz-b-000691430.
\end{enumerate}

\section{Related Work}
\subsection{Historical Map Segmentation}
Earlier approaches to historical map segmentation use conventional methods from computer vision such as Color Image Segmentation (CIS) and Template Matching (TM) which usually requires manual fine-tuning of parameters \cite{Chiang2014}. Nowadays, historical nap segmentation shift towards approaches based on machine learning or neural networks. CNN-based architectures, especially U-Net \cite{Ronneberger2015} and its variants, proved to be effective in segmenting various features from historical maps, such as building footprints \cite{Heitzler2020}, road networks \cite{Jiao2022}, hydrological features \cite{Wu2022}, and archaeological features \cite{Garcia-Molsosa2021}. More recently, Transformers \cite{Vaswani2017} have emerged as an alternative to CNNs for visual recognition \cite{Caron2021}. It is originally designed for Natural Language Processing (NLP), and has been later introduced to Computer Vision (CV) tasks and achieved promising performances due to its strong ability in attending and aggregating features on a global scale \cite{Dosovitskiy2020, Zheng2021}. However, Transformers are computationally more demanding and require more training data \cite{Caron2021}. To apply Transformers for historical map segmentation, \cite{Xia2023} designed a special contrastive pretraining strategy by leveraging image pairs depicting the same location while from different map series to create correlated views. Experimental results show that this contrastive pretraining strategy is effective in offsetting small labelled datasets on downstream segmentation tasks. 

Despite carefully designed pretraining strategy, available dataset from a single domain is still limited. Therefore, AI research is in a fast pace of paradigm shift from domain-specific CNNs or Transformers to foundation models trained on vast amount of data at scale \cite{Chen2023}. Research attempts lie with adapting a generalist into a specialist to fit the specific requirements from various domain. Successful practices have been witnessed in medical image segmentation \cite{Zhang2023, Wu2024}, remote sensing instance segmentation \cite{Chen2024}, license plate detection \cite{Ding2024}, and agricultural image segmentation \cite{Li2023} etc. However, this paradigm has not been introduced to the cartography domain yet. To the best of our knowledge, this pioneering work has been the first attempt to adapt a vision foundation model to historical map segmentation.

\subsection{Large Models in Segmentation}
The extraordinary scalability of Transformers has facilitated the development of large-scale models with billions of parameters. Initially, these models marked significant advancements in NLP, with examples such as BeRT \cite{Devlin2018} and LLaMA \cite{Touvron2023}, each demonstrating remarkable improvements over previous methods. More recently, attention has shifted towards large-scale vision models. These models are pre-trained on extensive masked data and exhibit strong generalization capabilities to new data distributions. A powerful foundation model for segmentation is Segment Anything Model (SAM) \cite{Kirillov2023}, which is trained on over 1 billion masks using a machine-in-the-loop data collection strategy. It can achieve zero-shot transfer to a range of downstream segmentation tasks via promotable segmentation. However, the generated masks of SAM have no labels. SEEM \cite{Zou2023} further incorporates semantic labels to the segmentation masks, and it allows for multi-modal prompt combinations, including visual, textual and audio. Painter \cite{Wang2023} and SegGPT \cite{Wang2023a} utilize a generalist in-context learning framework, which can segment any image by a given image-mask prompt indicating which task to perform. In this study, we choose the powerful yet elegant SAM model as our generalist model, aiming at customizing it to the historical map domain with enhanced performance.  
\begin{figure*}[!t]
\centering
\includegraphics[width=\textwidth]{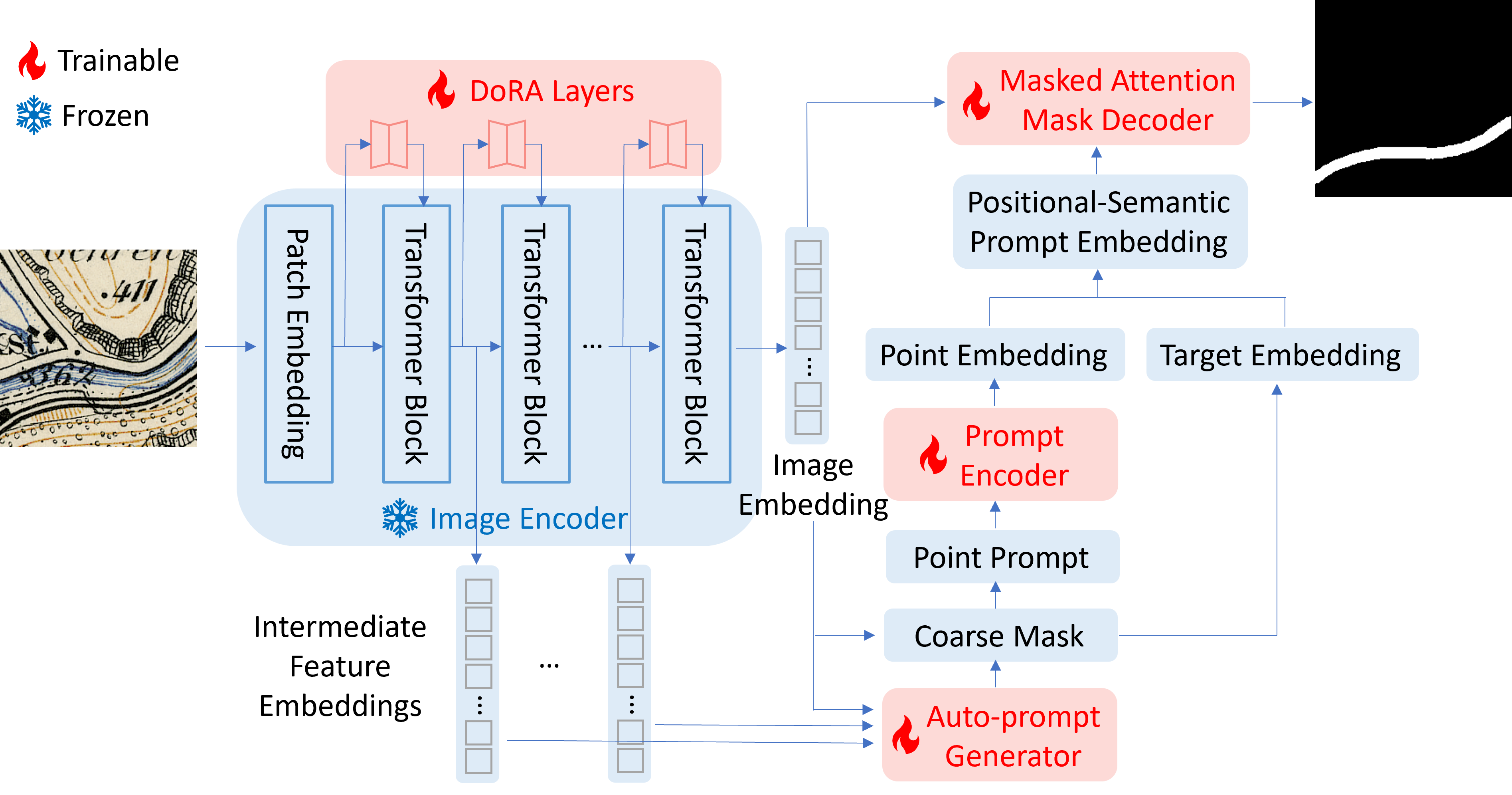}
\caption{The overall framework of MapSAM. We insert trainable DoRA layers into the image encoder to incorporate domain-specific feature information. The proposed auto-prompt generator leverages multi-layer feature embeddings to generate positive-negative point prompts. These prompts are combined with the target object's embedding to form high-level positional-semantic prompts. Finally, the positional-semantic prompts interact with the image embedding in the mask decoder using modified masked attention to generate the final object mask. }
\label{fig2}
\end{figure*}

\subsection{Parameter-Efficient Fine-Tuning}
Fine-tuning the full foundation model on downstream tasks can be impractical due to hardware memory and computational constraints. To address this issue, parameter-efficient fine-tuning (PEFT) is proposed, which aims to fine-tuning the foundation model by introducing only a small amount of trainable parameters while keeping the model backbone frozen \cite{Jia2022}. There are three main PEFT strategies, namely prompt tuning, adapter tuning and LoRA (Low-Rank Adaption) tuning. Prompt tuning \cite{Jia2022} modifies the input of the Transformer by prepending a set of learnable prompt embeddings into the original sequence of image patch embeddings. Adapter tuning \cite{Houlsby2019, Chen2022} adds additional residual blocks (or adapters) consisting of lightweight MLPs to the backbone Transformer layer. These two approaches utilize external embeddings or modules to incorporate task-specific information. LoRA tuning \cite{Hu2022} adopts a different strategy by constraining the update of a pre-trained weight matrix with a low-rank decomposition, which significantly reduces the number of trainable parameters required for downstream tasks. DoRA \cite{Liu2024} further decomposes the pre-trained weight into magnitude and direction components for fine-tuning, achieving even better performance and training stability than LoRA. We apply a state-of-the-art DoRA tuning strategy for adapting SAM to our specific task.

\section{Method}
Fig. \ref{fig2} illustrates the overall framework of our proposed MapSAM. When transferring SAM to historical map segmentation tasks by fine-tuning it with a domain-specific dataset, we freeze the heavyweight image encoder and keep the lightweight prompt encoder and mask decoder trainable. To enhance historical map feature extraction, we incorporate DoRA layers into the image encoder. Additionally, we designed an automatic prompt generator to create positive and negative point prompts. Combined with the embedding of the target object, this forms a positional-semantic prompt. The image embeddings are then efficiently queried by the positional-semantic prompt using modified masked attention within the mask decoder to produce object masks. The details of MapSAM are described in the following section.  

\subsection{Weight-Decomposed Low-Rank Adaptation for Image Encoder}
\begin{figure*}[!t]
\centering
\includegraphics[width=\textwidth]{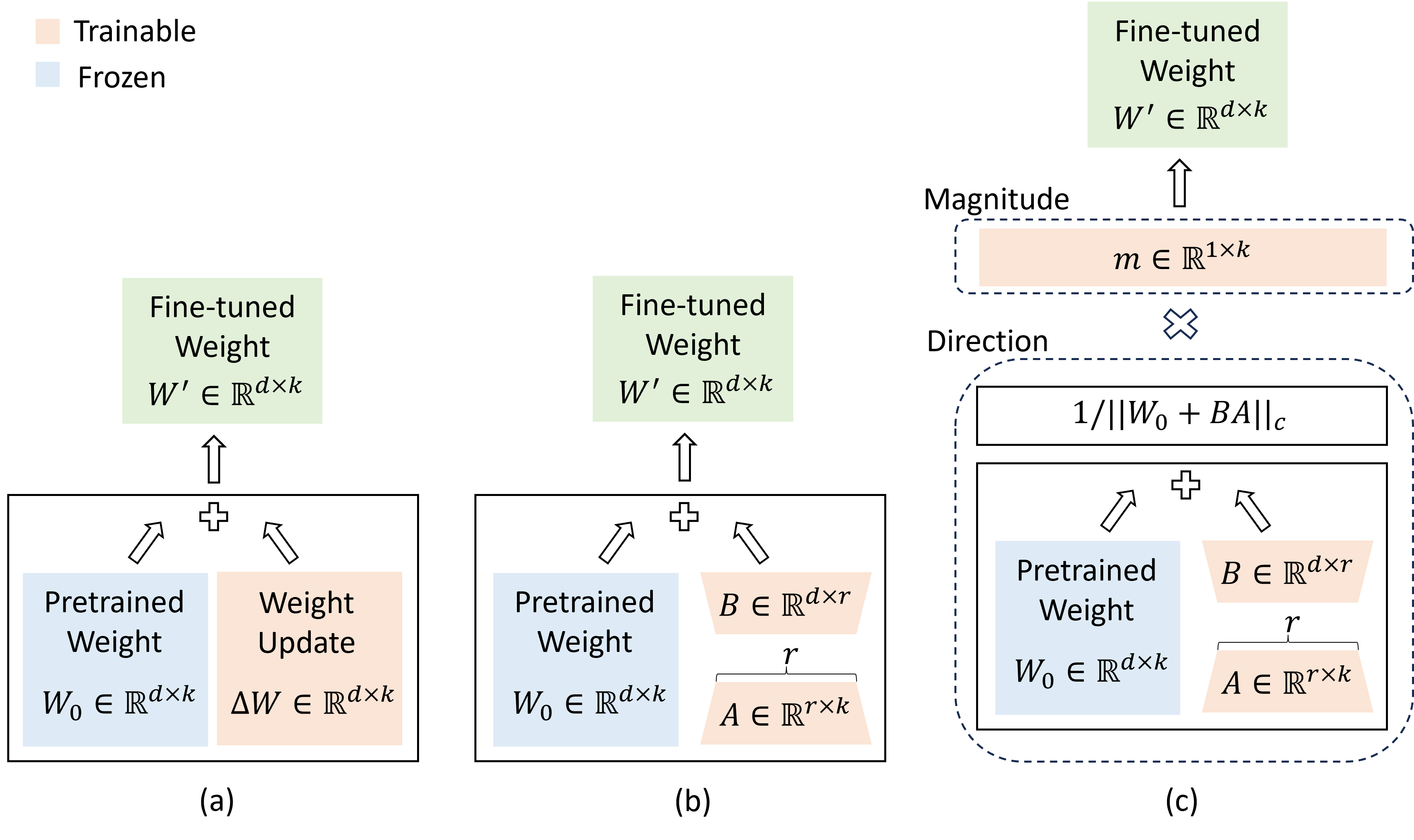}
\caption{Weight update mechanisms in (a) regular, (b) LoRA and (c) DoRA fine-tuning. In regular fine-tuning, the weight update matrix $\Delta W$ has the same dimensions as the pre-trained weight matrix. LoRA reduces the number of learnable parameters by approximating the weight update using two low-rank matrices, $B$ and $A$. DoRA further decomposes the weight update into magnitude and direction components, updating them separately: the magnitude matrix $m$ is trained directly, while the directional component $V$ is updated following the LoRA strategy. }
\label{fig3}
\end{figure*}
Low-Rank Adaptation (LoRA) \cite{Hu2022} reduces the number of trainable parameters by learning pairs of rank-decomposition matrices while keeping the original weights unchanged. This method significantly reduces the storage requirements for large-scale models tailored to specific tasks and facilitates efficient task-switching during deployment without increasing inference latency. The detail architecture of LoRA layers is illustrated in Fig. \ref{fig3}(b). For a pre-trained weight matrix $W_0 \in \mathbb{R}^{d \times k}$, the weight update in regular training and fine-tuning is defined as $W_0 + \Delta W$. The LoRA method provides a more efficient alternative for computing the weight updates $\Delta W$ by learning a low-rank decomposition, expressed as $BA$, where $B \in \mathbb{R}^{d \times r}$ and $A \in \mathbb{R}^{r \times k}$ are two low-rank matrices, with the rank $r \ll \min(d, k)$. Consequently, the fine-tuned weight $W'$ can be represented as:
\begin{equation}
W' = W_0 + \Delta W = W_0 + BA
\end{equation}

During fine-tuning, the pre-trained weight $W_0$ is frozen and does not receive gradient updates. Instead, the matrices $A$ and $B$, which contain trainable parameters, serve as a bypass to achieve the low-rank approximation for the weight update. 

DoRA (Weight-Decomposed Low-Rank Adaptation) \cite{Liu2024} builds on top of LoRA. Its architecture is illustrated in Fig. \ref{fig3}(c). DoRA restructures the pre-trained weight matrix $W_0 \in \mathbb{R}^{d \times k}$ into two separate components: a magnitude vector $m \in \mathbb{R}^{1 \times k}$ and a directional matrix $V \in \mathbb{R}^{d \times k}$. As shown in Eq. \ref{eq2}, with $\|V\|_c$ being the vector-wise norm of a matrix across each column, $V / \|V\|_c$ remains a unit vector, and $m$ defines the magnitude. During fine-tuning, the DoRA method applies LoRA to the directional component $V$ while allowing the magnitude component $m$ to be trained separately. As formulated in Eq. \ref{eq3}, the magnitude vector $m$ is a trainable vector, and $\Delta V$ represents the directional update learned through low-rank approximation. 
\begin{align}
\label{eq2}
W_0 &= m \frac{V}{\|V\|_c} \\
\label{eq3}
W' &= m \frac{V + \Delta V}{\|V + \Delta V\|_c} = m \frac{W_0 + BA}{\|W_0 + BA\|_c}
\end{align}

If maintaining the same rank, DoRA introduces a marginal increase of 0.01\% in trainable parameters compared to LoRA, due to the addition of the magnitude vector $m$ (a parameter of size $1 \times k$). However, across both LLM and vision transformer benchmarks, DoRA consistently outperforms LoRA, even when the DoRA rank is halved \cite{Liu2024}. Therefore, we adopt DoRA tuning in MapSAM.

The image encoder of SAM uses a ViT \cite{Ding2024} backbone, which primarily relies on multi-head self-attention \cite{Vaswani2017}. In multi-head self-attention, the input tokens are linearly projected into vectors called queries, keys and values. The output is determined by taking a weighted sum of the values, with the weights for each value calculated based on a compatibility function between the query and the corresponding key \cite{Vaswani2017}. In MapSAM, we inject DoRA layers into the projection layers of the queries and values to influence the output attention scores. Thus, the processing strategy of multi-head self-attention in MapSAM becomes:
\begin{align}
Q = W_q' \cdot x &= m_q \frac{W_q + B_q A_q}{\|W_q + B_q A_q\|_c} \cdot x \\
K &= W_k \cdot x \\
V = W_v' \cdot x &= m_v \frac{W_v + B_v A_v}{\|W_v + B_v A_v\|_c} \cdot x \\
\text{Attention}(Q, K, V) &= \text{softmax}\left(\frac{Q K^T}{\sqrt{d}}\right) V
\end{align}
where $W_q$, $W_k$, and $W_v$ are the pre-trained weight matrices from SAM, $m_q$, $m_v$, $A_q$, $B_q$, $A_v$, and $B_v$ are the trainable DoRA parameters, and $x$ represents the input image tokens.   
\begin{figure*}[!t]
\centering
\includegraphics[width=\textwidth]{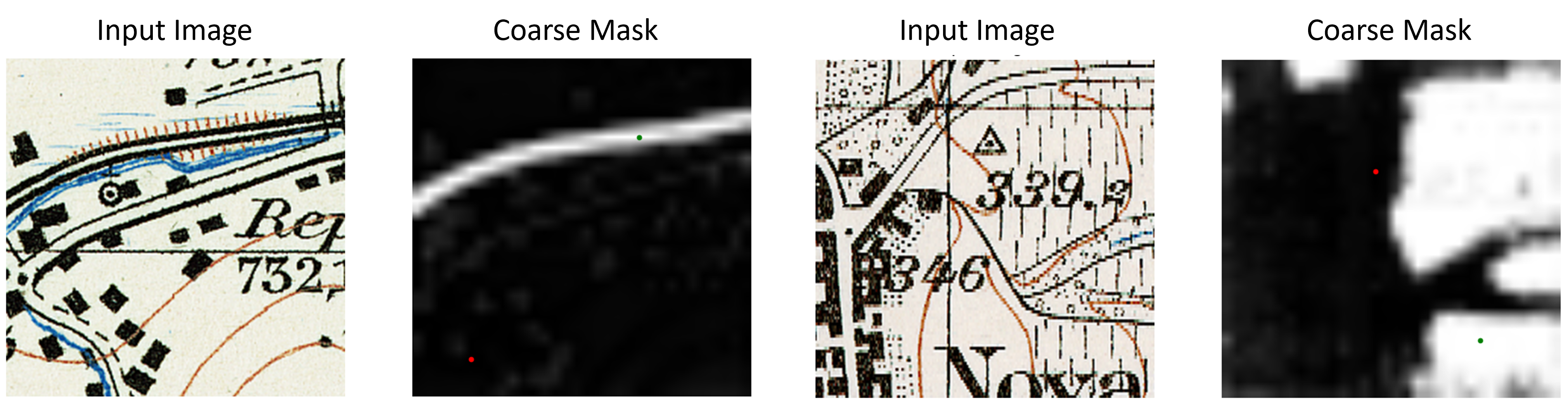}
\caption{The coarse mask generated by the auto-prompt generator, with the corresponding positive (green) and negative (red) point prompts, applied to input images for railway detection (left) and vineyard detection (right).}
\label{fig4}
\end{figure*}

\subsection{Automatic Positional-Semantic Prompting}
The original SAM performs "promptable segmentation", which requires manual annotations such as points or bounding boxes to serve as prompts. These prompts provide the model with additional context for segmentation by indicating where to focus. In practice, manually labeling points or bounding boxes for thousands or even millions of historical map images is extremely expensive and time-consuming. To address this issue, we propose an auto-prompt generator that leverages multi-layer feature information obtained from the image encoder to generate prompts automatically. 

Let $I \in \mathbb{R}^{W \times H \times 3}$ be the input image, the image encoder of SAM generates 16× downscaled feature embeddings $F \in \mathbb{R}^{\frac{W}{16} \times \frac{H}{16} \times 256}$ of the input image. As depicted in Fig. \ref{fig2}, the auto-prompt generator takes these feature embeddings as input to predict a binary coarse mask $M \in \mathbb{R}^{\frac{W}{16} \times \frac{H}{16} \times 1}$ of the foreground regions. To reduce dimensionality, the auto-prompt generator utilizes 3 layers of $1 \times 1$ convolutions with LayerNorm layers \cite{Ba2016} on the feature embeddings. Existing research has shown that using a pretrained network’s intermediate representations can significantly improve generalization ability on out-of-domain tasks \cite{Evci2022}. Therefore, we combine features from various encoder layers to predict the coarse mask $M$. The feature selection process is further explained in the ablation study presented in Section \ref{sec4}. 

The selected multi-layer features are fed into the auto-prompt generator to produce a sequence of foreground predictions. We then take the average of these predictions as a simple yet effective way to fuse multi-layer information, generating the final binary coarse mask to guide the rough location of the target object. To generate positive and negative location priors as segmentation hints for prompting SAM, the predicted coarse mask is interpolated to match the original size of the input image. Positive and negative point prompts are then automatically generated by identifying the locations with the highest and lowest probability scores in the interpolated mask (Fig. \ref{fig4}).

In SAM, the point prompts are fed into the prompt encoder and converted into vectorial embeddings $P \in \mathbb{R}^{N_{\text{tokens}} \times 256}$, which contain the positional encoding of the point’s location and sparse semantic information indicating foreground or background. These prompt embeddings are then interacting with the image embedding in the mask decoder to generate the output mask. To provide SAM with additional visual cues in the prompt tokens, similar to PerSAM \cite{Zhang2023a}, we first extract the target object's embedding from the feature embedding $F$ using the coarse mask $M$. The final global embedding $T$ for the target object is obtained by averaging the local features within the target area through average pooling:
\begin{equation}
T = \frac{1}{n} \sum_{i=1}^{n} T_i \in \mathbb{R}^{1 \times 256}
\end{equation}
where $T_i$ represents the feature embedding of the target object at pixel $i$. We then element-wise add the target embedding $T$ to the point embedding $P$ to provide additional high-level semantic prompting.  

By incorporating point prompts that provide low-level positional information and target embeddings that offer high-level feature information, MapSAM achieves automatic positional-semantic prompting, eliminating the need for any manual intervention.

\subsection{Masked-Attention Mask Decoder}
\begin{figure*}[!t]
\centering
\includegraphics[width=\textwidth]{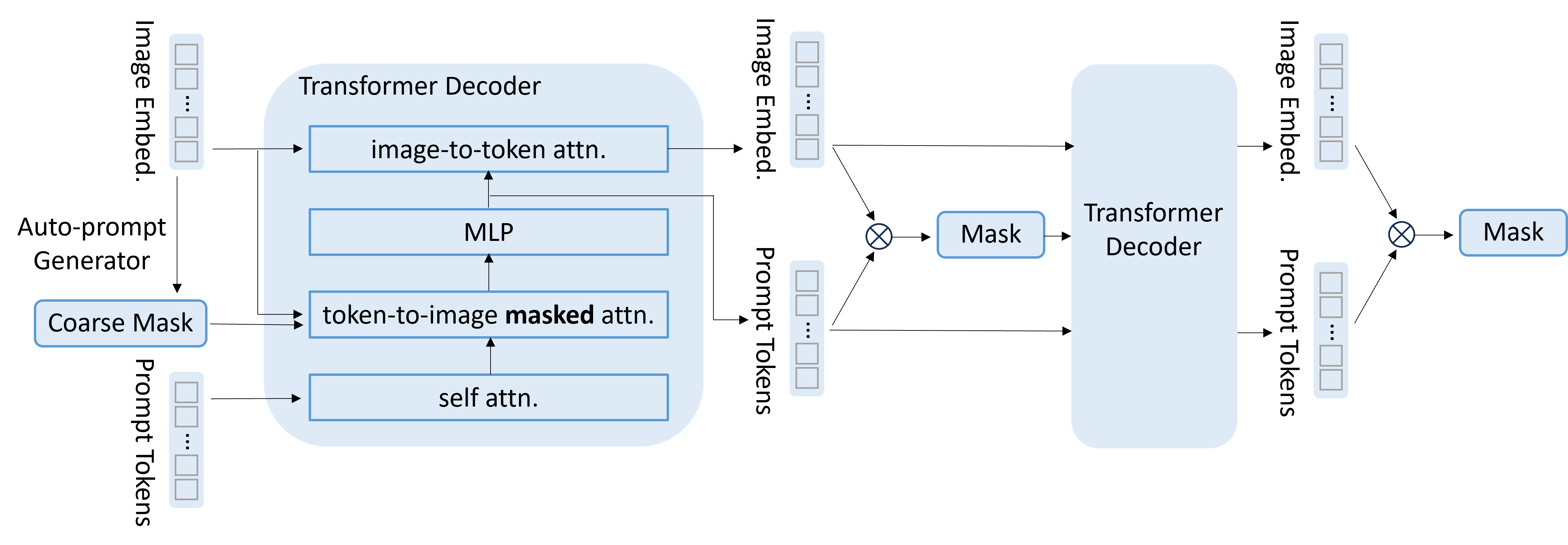}
\caption{The modified masked-attention mask decoder in MapSAM. The coarse mask produced by the auto-prompt generator serves as the initial mask for token-to-image masked attention. As the process progresses through each decoder layer, this attention mask is iteratively refined by incorporating the updated image embeddings and prompt tokens, allowing for more precise modulation and enhanced accuracy.}
\label{maskedattn}
\end{figure*}
The lightweight mask decoder in SAM efficiently maps the prompt tokens and the image embedding to the output mask. This is achieved through self-attention on the prompt tokens and bidirectional cross-attention, where prompt tokens serve as queries for the image embedding and vice versa. However, cross-attention layers are claimed to have slow convergence and difficulty in learning to attend to localized object regions due to their global context \cite{Cheng2022}. To address this, Mask2Former \cite{Cheng2022} proposes masked attention, compelling cross-attention to focus only on foreground target regions. 

To effectively capture the visual semantics of the target object and avoid focusing on irrelevant background areas, we enhance the token-to-image cross-attention layer in SAM by incorporating the masked attention mechanism. This mechanism adjusts the attention matrix using an attention mask $\mathcal{M}_{l-1}$:
\begin{equation}
X_l = \text{softmax}\left(\mathcal{M}_{l-1} + Q_l K_l^T\right) V_l + X_{l-1}
\end{equation}
Here, $l$ denotes the layer index, and $X_l$ represents the query features at the $l$-th layer. The term $Q_l = f_Q\left(X_{l-1}\right)$ refers to the transformed query features from the previous layer. $K_l$ and $V_l$ are the image features transformed by the functions $f_K(\cdot)$ and $f_V(\cdot)$, respectively. The functions $f_Q$, $f_K$, and $f_V$ are linear transformations.  For a feature location $(x,y)$, the attention mask is defined as:
\begin{equation}
\mathcal{M}_{l-1}(x, y) =
\begin{cases} 
    0, & \text{if } M_{l-1}(x, y) = 1 \\
    -\infty, & \text{otherwise}
\end{cases}
\end{equation}
where $M_{l-1}(x,y)$ is the binarized mask prediction of the previous Transformer decoder layer. We use the coarse mask predicted by the auto-prompt generator as the initial mask for the image features. After each decoder layer, the mask is refined using the binary mask prediction generated by a mask head. This mask head, similar to the one used in SAM, employs a spatially point-wise product between the updated tokens and the image embeddings to predict the mask. This process is illustrated in Fig. \ref{maskedattn}. With the modified masked attention in the mask decoder, MapSAM directs the prompt tokens to focus primarily on foreground target regions, facilitating more effective feature aggregation and enhancing the final segmentation accuracy.
\begin{table*}[h]
    \centering
    \caption{Comparison of the proposed MapSAM with other baselines on the Railway and Vineyard datasets. The best results of IoU and F1 are in \textbf{bold}.}
    \label{tab1}
    \setlength{\extrarowheight}{0.1mm}
    \setlength\tabcolsep{5pt}
    \resizebox{1.00\textwidth}{!}{
        \begin{tabular}{c|cccccccc|cccc}
            \hline
            \hline
            \multirow{3}{*}{Method} 
            & \multicolumn{8}{c|}{Railway}                                                                                                         
            & \multicolumn{4}{c}{Vineyard}                                   \\ 
            
            \cline{2-13} 
            & \multicolumn{2}{c}{Full (5872)} & \multicolumn{2}{c}{10\% (587)} & \multicolumn{2}{c}{1\% (58)} & \multicolumn{2}{c|}{10-shot} 
            & \multicolumn{2}{c}{Full (613)} & \multicolumn{2}{c}{10-shot} \\

            \cline{2-13}	
            & F1 & IoU & F1 & IoU & F1 & IoU & F1 & IoU & F1 & IoU & F1 & IoU \\
            \hline

            U-Net
            & \textbf{95.13} & \textbf{91.86} & \textbf{94.30} & \textbf{90.56} 
            & 89.68 & 83.52 & 74.02 & 61.43 
            & \textbf{84.61} & \textbf{77.04} & 71.52 & 60.23 \\

            SAMed 
            & 91.98 & 86.31 & 91.63 & 85.69 
            & 91.75 & 86.01 & 84.62 & 75.44 
            & 82.80 & 74.85 & \textbf{71.95} & \textbf{61.53} \\

            Few-Shot SAM 
            & -- & -- & -- & -- 
            & -- & -- & 47.49 & 35.82 
            & -- & -- & 57.99 & 46.83 \\

            MapSAM
            & 94.06 & 89.46 & 93.57 & 88.71 
            & \textbf{92.05} & \textbf{86.53} & \textbf{87.17} & \textbf{78.50} 
            & 82.78 & 74.32 & 70.52 & 60.02 \\ 
            \hline
            \hline
        \end{tabular}
    }
\end{table*}

\section{Experiment}
\subsection{Datasets}
We apply MapSAM to two datasets to evaluate its performance in segmenting different features from historical maps: one railway dataset representing linear features and one vineyard dataset representing areal features. Both datasets are manually labeled and tiled into a fixed size of $224 \times 224$ pixels from the Swiss Siegfried Maps. Each dataset contains binary segmentation logits, including one background class and one target class (railway or vineyard).

The full railway dataset contains 5,872 training tiles, 839 validation tiles, and 1,679 testing tiles, following a rough splitting ratio of 7:1:2. To explore the resources needed to effectively fine-tune a foundation model for the downstream segmentation task, we progressively reduce the number of labeled training data to 10\% (587 tiles), 1\% (58 tiles), and even to just 10 tiles (10-shot). We then evaluate the model’s performance on the entire testing set in each case. The vineyard dataset is smaller, containing 613 training tiles, 87 validation tiles, and 177 testing tiles. We also tested the 10-shot training on the vineyard dataset.

\subsection{Implementation Details}
The utilized base model is the “ViT-B” version of SAM. The output resolution of the predicted segmentation logit is 4× downscaled of the input image size. To maintain a reasonably sized segmentation logit, we upsampled the $224 \times 224$ input images to $512 \times 512$ before feeding them to MapSAM. The final output mask is interpolated back to the original size of the input image. Unlike SAM’s ambiguous predictions, MapSAM deterministically predicts a single binary mask for each semantic class.

The overall loss of MapSAM consists of two parts: the segmentation loss of the coarse mask predicted by the auto-prompt generator and the loss of the final segmentation mask. We supervise both masks using a weighted combination of focal loss $L_{\text{focal}}$ \cite{Lin2020} and dice loss $L_{\text{dice}}$ \cite{Sudre2017} as follows:
\begin{align}
L_{\text{coarse}} &= \lambda L_{\text{focal}} (\tilde{x}_i^{\text{coarse}}, x_i^{\text{gt}}) + (1 - \lambda) L_{\text{dice}} (\tilde{x}_i^{\text{coarse}}, x_i^{\text{gt}}) \\
L_{\text{final}} &= \lambda L_{\text{focal}} (\tilde{x}_i^{\text{final}}, x_i^{\text{gt}}) + (1 - \lambda) L_{\text{dice}} (\tilde{x}_i^{\text{final}}, x_i^{\text{gt}}) \\
L_{\text{overall}} &= L_{\text{coarse}} + L_{\text{final}}
\end{align}
where $\tilde{x}_i^{\text{coarse}}$ is the predicted segmentation probability for pixel $i$ in the coarse mask, and $\tilde{x}_i^{\text{final}}$ denotes the predicted probability in the final segmentation mask. $x_i^{\text{gt}}$ represents the corresponding ground truth for pixel $i$. $\lambda$ is the coefficient to balance the two loss terms, which is set to 0.2 in our case. 

The model performance is evaluated using the F1 score and Intersection over Union (IoU), which are commonly used metrics for dense predictions. Our experiments were implemented in PyTorch 1.9.1 on an Nvidia Quadro RTX 5000 16 GB GPU.  We utilized a batch size of 8, set the base learning rate to 0.005, and trained MapSAM for 160 epochs. The AdamW optimizer was used to optimize the training process. To stabilize training, we adopted the same Warmup strategy \cite{He2016} as SAMed \cite{Zhang2023}. This strategy involves gradually increasing the learning rate from a lower initial value to the base learning rate at the beginning of training, allowing for healthy convergence and avoiding issues associated with high learning rates in the initial stages. After the warmup period, the learning rate is controlled by exponential decay to ensure gradual convergence. The learning rate adjustment strategy can be summarized as follows:
\begin{equation}
\text{lr} = 
\begin{cases} 
\frac{t \times B_{\text{lr}}}{T_w} & \text{for } t \leq T_w \\
B_{\text{lr}} \left(1 - \frac{t - T_w}{T_{\text{max}}} \right)^{0.9} & \text{for } t > T_w 
\end{cases}
\end{equation}
Where $t$ is the current training iteration, $T_w$ is the warmup period (the number of iterations for warmup), $T_{\text{max}}$ is the maximum number of iterations, and $B_{\text{lr}}$ is the base learning rate.  

\subsection{Experimental Results}
We compare the performance of MapSAM on two downstream tasks—railway detection and vineyard detection—against commonly used historical map segmentation models and advanced large models with adaptation techniques customized for the medical image domain. For commonly used historical map segmentation models, we use U-Net as a representative. For baselines using large models with adaptation strategies, we choose SAMed \cite{Zhang2023} and Few-Shot SAM \cite{Wu2024}. SAMed employs the LoRA fine-tuning strategy and achieves state-of-the-art performance on the Synapse multi-organ segmentation dataset, while Few-Shot SAM incorporates a self-prompt unit and demonstrates strong performance in few-shot settings. 
\begin{figure*}[t]
\centering
\includegraphics[width=\textwidth]{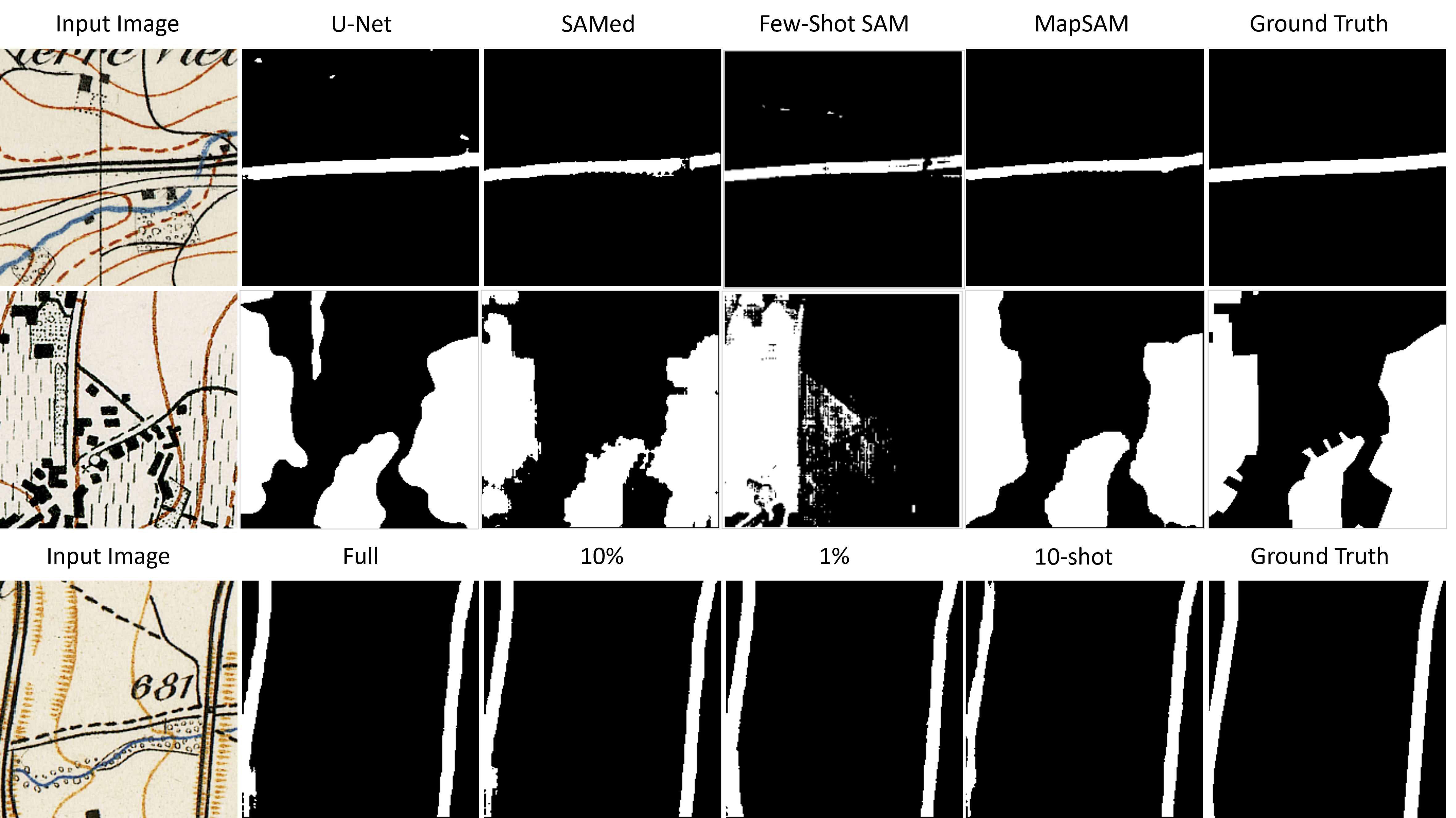}
\caption{The top two rows display the performance of different models under the 10-shot setting for railway and vineyard detection. The bottom row shows MapSAM's performance in railway detection when trained on the full dataset, as well as on 10\%, 1\%, and 10-shot subsets.}
\label{fig5}
\end{figure*}

According to Table \ref{tab1}, when sufficient training data are available, domain-specific models like U-Net continue to demonstrate superior performance compared to fine-tuned foundation models. However, with limited training data, fine-tuned foundation models emerge as a strong alternative for downstream segmentation tasks. The proposed MapSAM outperformed SAMed and Few-Shot SAM in detecting linear features (e.g., railways) in historical maps, though it performed slightly worse than SAMed in detecting areal features (e.g., vineyards). SAMed achieves good performance in historical map segmentation due to its injection of LoRA layers into the image encoder, allowing it to incorporate domain-specific knowledge. In contrast, Few-Shot SAM, which only trains the self-prompt unit while keeping the image encoder frozen, exhibits weaker generalization to historical map data. Overall, MapSAM demonstrates a promising ability to adapt foundation models for downstream segmentation tasks in the historical map domain, showing strong generalization across various historical map features without requiring structural changes. 

The results are visualized in Fig. \ref{fig5}. These visualizations demonstrate that MapSAM effectively adapts the generalist SAM into a specialist for historical map segmentation with minimal effort, such as fine-tuning with just 10 examples. MapSAM meets domain-specific requirements, such as maintaining continuity and clear boundaries for delineated objects, particularly for challenging linear features that extend over large areas with narrow widths. In contrast, other baseline models show more errors or omissions (e.g., U-Net, Few-Shot SAM) or suffer from discontinuities and blurred boundaries (e.g., SAMed, Few-Shot SAM). The superior segmentation performance of MapSAM is attributed to DoRA tuning, which enhances the model's segmentation accuracy, as well as the higher-level positional-semantic prompting and masked attention, which refine segmentation shapes. Furthermore, MapSAM exhibits stable performance even when the amount of fine-tuning data is significantly reduced, from the full dataset down to just 10 examples. Despite this reduction, MapSAM continues to accurately detect target railway objects while preserving geometric continuity.

With its consistently strong performance across various map features, high-quality geometric outputs, low-resource adaptability, and fully automated capabilities, MapSAM opens up new possibilities for historical map segmentation.

\subsection{Ablation Study}
\subsubsection{Main Components}
We test the effectiveness of the main components in MapSAM by starting with a basic version and then incrementally adding other components. The most basic variant of MapSAM is equipped only with the auto-prompt generator, which achieves automatic historical map segmentation without human intervention. We conduct ablation studies on the partial railway dataset with 1\% training data. As shown in Table \ref{tab2}, MapSAM with only automatic point prompts achieves an IoU of 70.91\%. Adding DoRA significantly enhances performance, improving the IoU by 14.25\%, which fully indicates the importance of incorporating domain-specific knowledge in the image encoder. Building on the positional prompt, we introduce high-level target semantic prompting, resulting in an additional 0.72\% improvement in IoU. Finally, we use masked attention for more effective feature aggregation, which further boosts the IoU score by 0.65\%. 
\begin{table}[]
    \centering
    \caption{Ablation of main components in MapSAM.}
    \label{tab2}
    \setlength{\extrarowheight}{0.1mm}
    \setlength\tabcolsep{5pt}
    \resizebox{0.7\columnwidth}{!}{
        \begin{tabular}{lcc}
            \hline
            \hline
            \textbf{MapSAM Variant} & \textbf{IoU} & \textbf{Gain} \\
            \hline
            Positional Prompt  & 70.91 & -      \\
            + DoRA             & 85.16 & +14.25 \\
            + Semantic Prompt  & 85.88 & +0.72  \\
            + Masked Attention & 86.53 & +0.65  \\
            \hline
            \hline
        \end{tabular}
    }
\end{table}

\subsubsection{Multi-layer feature selection}
\label{sec4}
The coarse mask in MapSAM serves as a crucial intermediate output. It is used to generate the point prompt, identify the foreground region for extracting the target embedding, and act as the initial mask for masked attention. To create an efficient feature representation for predicting the coarse mask, we combine features from different encoder layers. A key hyperparameter to tune is the selection of encoder layers from which to extract feature embeddings. To investigate the optimal multi-layer feature selection, we conducted comprehensive ablation studies, as shown in Table \ref{tab3}. Since we use ViT-B with 12 Transformer encoder layers as our backbone, we have 12 feature embedding layers to choose from and integrate.
\begin{table}[]
    \centering
    \caption{Ablation of multi-layer feature selection.}
    \label{tab3}
    \setlength{\extrarowheight}{0.1mm}
    \setlength\tabcolsep{15pt}
    \resizebox{0.9\columnwidth}{!}{
        \begin{tabular}{lc}
            \hline
            \hline
            \textbf{Feature Embedding Layers} & \textbf{IoU} \\
            \hline
            1 layer \{12\}             & 84.10          \\
            2 layers \{6,12\}          & 85.19          \\
            3 layers \{4,8,12\}        & 86.04          \\
            4 layers \{3,6,9,12\}      & 84.30          \\
            4 layers \{1,5,9,12\}      & \textbf{86.14} \\
            6 layers \{2,4,6,8,10,12\} & 84.84          \\
            \hline
            \hline
        \end{tabular}
    }
\end{table}

According to the quantitative results in Table \ref{tab3}, integrating features from multiple layers consistently improves the performance of our entire pipeline compared to using only the last layer's feature embedding without multi-layer fusion. The optimal combination of feature embeddings for fusion is the sequence \{1, 5, 9, 12\}. 

\begin{figure*}[!t]
\centering
\includegraphics[width=\textwidth]{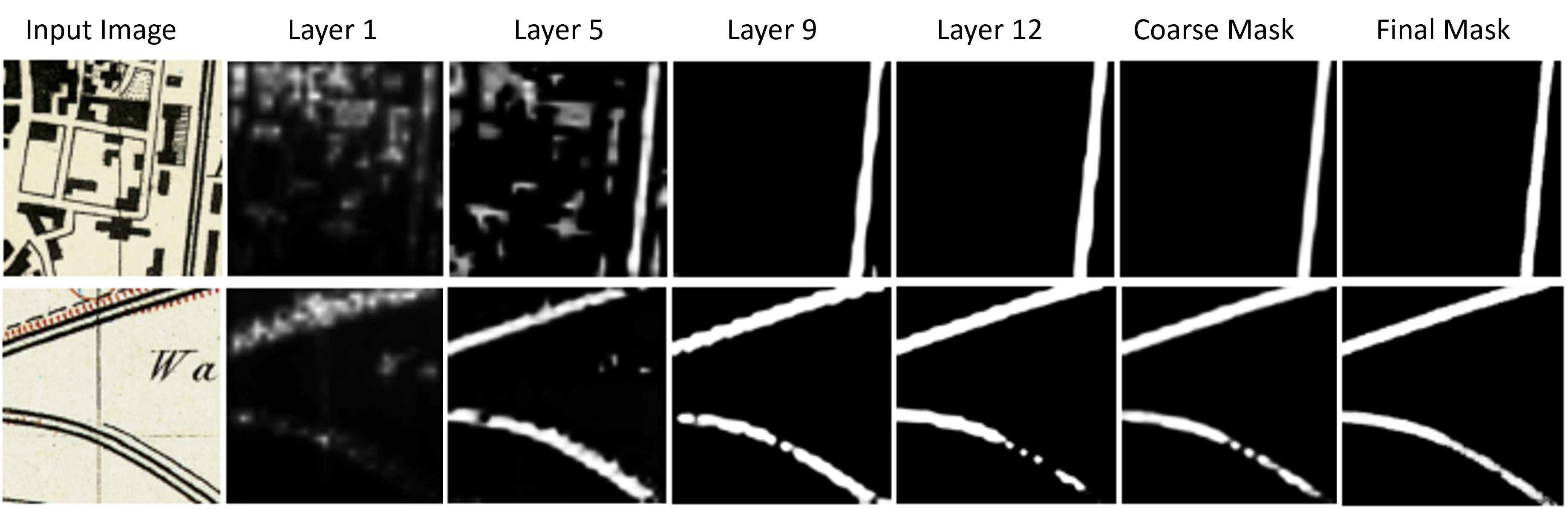}
\caption{The intermediate mask predicted by different feature embedding layers, the fused coarse mask and the final segmentation mask.}
\label{fig6}
\end{figure*}

To provide an intuitive understanding of how different feature embedding layers contribute to the final coarse mask, we visualized the intermediate mask outputs generated by feeding various feature embedding layers into the auto-prompt generator. Each visualization was obtained by applying interpolation and a sigmoid function. As shown in Fig. \ref{fig6}, the earlier layers contribute to more general feature extraction, while the later layers capture more task-specific features. After multi-layer fusion, a significant improvement in pixel-level accuracy along boundary lines is observed. Moreover, compared to the coarse mask, the final segmented results inferred from the positional-semantic prompt show further improvement, highlighting the importance of the interaction between prompt tokens and feature embeddings within the mask decoder using the modified masked attention mechanism.

\section{Conclusion}
In this paper, we introduce MapSAM, a customized adaptation of the visual foundation model SAM for the historical map domain. To better incorporate domain-specific knowledge, we insert trainable DoRA layers into the frozen image encoder of SAM. We enhance the model by integrating an automated positional-semantic prompt generation process, which prompts SAM with both low-level positional information and high-level target semantics, eliminating the need for human intervention. Additionally, we modify the cross-attention mechanism in the mask decoder with masked attention for more intensive feature aggregation. We evaluated the proposed MapSAM on two downstream tasks focused on segmenting different historical map features and achieved promising results. Our findings suggest that efficient fine-tuning of visual foundation models offers a new approach to historical map segmentation, particularly when training data is limited.

\bibliographystyle{IEEEtran}

\vfill
\end{document}